%% file: main.tex
\documentclass[10pt,twocolumn,letterpaper]{article}

\usepackage[pagenumbers]{cvpr} 
\input{preamble}
\definecolor{cvprblue}{rgb}{0.21,0.49,0.74}
\usepackage[pagebackref,breaklinks,colorlinks,allcolors=cvprblue]{hyperref}
\usepackage{algorithm}
\usepackage{algorithmic}
\usepackage{amsmath} 
\usepackage{amsfonts}
\usepackage{multirow}
\usepackage{colortbl}
\def\confName{CVPR}
\def\confYear{2027}

\title{Spatio-Temporal Adversarial Attack for Asynchronous Event Streams}

\author{Hongwei Ren\\
Harbin Institute of Technology\\
\and
Lansong Yang\\
Harbin Institute of Technology\\
\and
Youxin Jiang\\
Harbin Institute of Technology\\
\and
Tuopusen Huang\\
Harbin Institute of Technology\\
\and
Xiangqian Wu\\
Harbin Institute of Technology\\
}

\begin{document}
\maketitle
\input{sec/0_abstract}    
\input{sec/1_intro}

\input{sec/2_releated_work}
\input{sec/3_method}
\input{sec/4_experiment}
{
    \small
    \bibliographystyle{ieeenat_fullname}
    \bibliography{main}
}


\end{document}

%% file: sec/0_abstract.tex
\begin{abstract}
Event cameras capture sparse and asynchronous visual signals with high temporal resolution and low latency, enabling emerging applications such as robotics and autonomous systems. However, event-based recognition models remain vulnerable to adversarial perturbations, while adversarial attacks on asynchronous event streams are still largely unexplored. This is partly because many existing event representations rely on discretization or aggregation operations, making direct optimization over individual events difficult. In this paper, we propose a spatio-temporal adversarial framework for generating adversarial event streams. Different from geometry-oriented point cloud attacks, our method explicitly exploits the spatial and temporal structures of asynchronous events during adversarial optimization. Specifically, we introduce a spatio-temporal perturbation diffusion strategy that propagates perturbations over spatial and temporal event neighborhoods, together with a sample-wise adaptive optimization scheme to improve attack reliability. Extensive experiments on multiple event recognition benchmarks and backbones demonstrate that our method achieves 100\% attack success rates with substantially reduced perturbation costs. Furthermore, event-specific evaluations and defense experiments show that the generated adversarial events maintain lower spatio-temporal distortion while preserving strong attack effectiveness.
\end{abstract}

%% file: sec/1_intro.tex
\section{Introduction}
\label{sec:intro}
\begin{figure}
\includegraphics[width=0.9\linewidth]{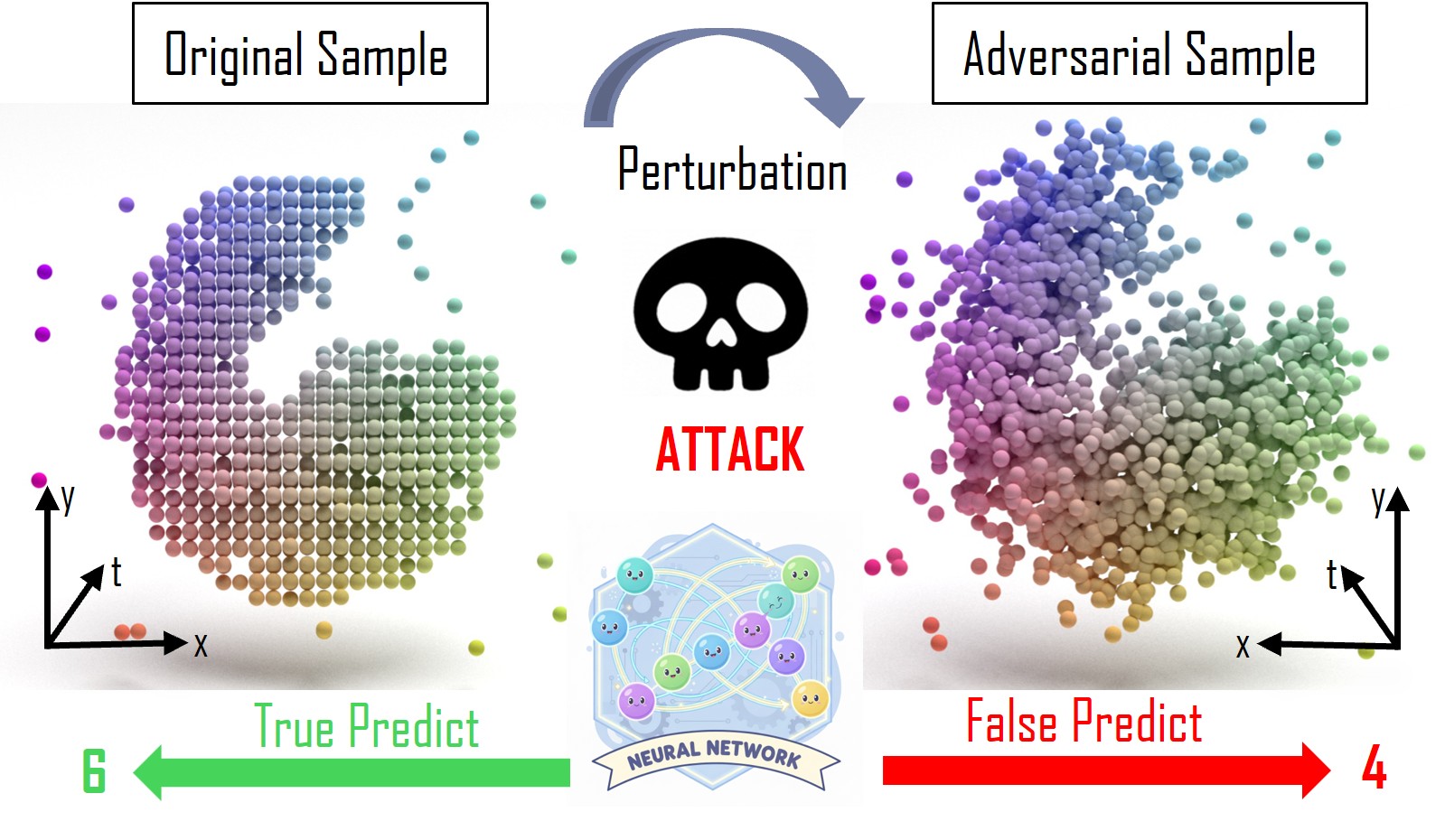}
\caption{A representative adversarial example from the N-MNIST. After the attack method adds perturbations to events, the network misclassifies the handwritten digit 6 as 4.}
\label{fig: adversarial sample}
\end{figure}

Deep Neural Networks (DNNs) have demonstrated remarkable performance across a wide range of tasks in Computer Vision (CV) and Natural Language Processing (NLP) \cite{lecun2015deep}.
However, DNNs are also known to be vulnerable to adversarial examples, where carefully crafted perturbations to the input can induce incorrect predictions \cite{xiang2019generating}, as illustrated in \cref{fig: adversarial sample}. 
Such vulnerabilities raise concerns for vision systems deployed in applications such as autonomous systems and robotics.
Adversarial robustness has therefore been extensively studied for conventional modalities, including 2D images, text, and 3D point clouds \cite{zhang2025adversarial}. In contrast, adversarial attacks on event-based recognition remain relatively underexplored, particularly for models that directly process sparse and asynchronous event streams.

Event cameras, inspired by biological vision systems, provide an asynchronous sensing paradigm by recording pixel-level brightness changes with microsecond-level temporal resolution \cite{gallego2020event}. Unlike conventional cameras that sample intensity at fixed time intervals, event cameras respond to changes in the scene and produce sparse event streams with high temporal resolution, high dynamic range, and low latency. These properties make them particularly suitable for applications involving rapid motion, such as high-speed tracking and real-time robotics.
Many existing event-based models first transform asynchronous events into aggregated representations, such as event frames or voxel grids, before subsequent processing \cite{zhu2018ev}.
While effective for recognition, such aggregation alters the direct correspondence between individual events and model inputs, making it difficult to explicitly optimize and evaluate perturbations at the event level. 
This motivates studying adversarial perturbations directly in the asynchronous event domain.

Point-based representations have recently emerged as an effective paradigm for processing asynchronous event streams, as they preserve individual events together with their spatial and temporal information. Point-based architectures such as PointNet++ \cite{qi2017pointnet++}, together with event-specific models such as EventMamba \cite{ren2025rethinking}, have demonstrated promising performance for event recognition \cite{ren2024spikepoint}. 
Meanwhile, adversarial attacks on conventional point clouds have been extensively studied, where perturbations are typically optimized according to geometric structures and spatial distributions \cite{kurakin2018adversarial,xiang2019generating}.
However, although point clouds and event streams can both be represented as sets of points, their underlying information is fundamentally different: \textbf{conventional point clouds primarily encode static geometric structure, whereas event streams additionally contain asynchronous temporal variations}. Consequently, directly applying geometry-oriented point cloud attacks may fail to account for the temporal dependencies and local spatio-temporal structures inherent in event data. This motivates adversarial event generation that explicitly considers both spatial and temporal relationships during perturbation optimization. 

In this paper, we present ST-ADV, a spatio-temporal adversarial framework for generating adversarial event streams on point-based event recognition models. Unlike conventional point cloud attacks that mainly optimize geometric perturbations, our method explicitly considers the spatial and temporal structures of asynchronous events during adversarial optimization. Specifically, we introduce a spatio-temporal perturbation diffusion strategy that propagates perturbations over spatial and temporal event neighborhoods, encouraging locally consistent perturbations over sparse asynchronous streams. Furthermore, we adopt a sample-wise adaptive optimization strategy to improve optimization reliability by accounting for different attack difficulties across individual samples. Our goal is not merely to generate effective adversarial examples, but also to provide a systematic framework for evaluating the adversarial robustness of point-based event recognition models.
Our contributions can be summarized as follows:
\begin{itemize}
    \item We formulate adversarial event generation as a direct optimization problem over asynchronous event streams and develop a gradient-based attack framework for point-based event recognition models.
    
    \item We introduce a spatio-temporal perturbation diffusion strategy that propagates perturbations over spatial and temporal event neighborhoods to encourage locally consistent adversarial modifications.
    
    \item We incorporate a sample-wise adaptive optimization strategy as an optimization refinement for handling samples with different attack difficulties.
    
    \item Extensive experiments demonstrate high attack success rates with substantially reduced geometric and temporal distortions, revealing the vulnerability of point-based event recognition models.
\end{itemize}

%% file: sec/2_releated_work.tex
\section{Related Work}
In this section, we review representative adversarial attack methods for point clouds and event-based vision, with particular emphasis on their perturbation spaces and structural constraints for irregular inputs.

\subsection{Point Cloud Adversarial Attack}
A variety of adversarial attacks have been developed for point cloud recognition models. Early approaches extended gradient-based attacks such as FGSM and C\&W to point sets by perturbing point coordinates or manipulating point distributions \cite{liu2019extending,xiang2019generating}. Subsequent studies improved attack imperceptibility and geometric consistency through geometry-aware constraints, shape-preserving perturbations, and structure-guided optimization, including Geo$\text{A}^{3}$ \cite{wen2020geometry}, SI-ADV \cite{huang2022shape}, Manifold Attack \cite{tang2023deep}, and HIT-ADV \cite{lou2024hide}.
Other studies have investigated transferable
adversarial perturbations~\cite{hamdi2020advpc}
and minimal point manipulations
~\cite{kim2021minimal} to improve attack
effectiveness while limiting perceptible
distortions.
More recently, physically realizable attacks such as SmokeAttack \cite{wei2025smokeattack} have explored adversarial perturbations beyond purely digital point manipulation. Despite these advances, existing point cloud attacks are designed to preserve or exploit static geometric structures, whereas asynchronous event streams additionally contain explicit temporal information and local temporal dependencies.

\subsection{Event Adversarial Attack}
Adversarial attacks for event-based vision remain comparatively underexplored. Early studies investigated event-specific adversarial generation through operations such as timestamp modification and event insertion or removal \cite{lee2022adversarial,buchel2022adversarial,yao2024exploring}. These approaches demonstrate that event-based models are vulnerable to adversarial manipulation, but they operate under perturbation spaces and processing pipelines that differ from continuous optimization over point-based event inputs. More recently, physical adversarial attacks have also been explored for event-based pedestrian detection by optimizing adversarial textures on physical carriers \cite{lin2025adversarial}. Such methods target the sensing or scene level and therefore address a different threat model from direct digital perturbation of asynchronous event inputs.

Overall, prior studies have established the vulnerability of both point-based models and event-based vision systems, but the direct optimization of adversarial perturbations over asynchronous event streams remains insufficiently studied. In particular, conventional point cloud attacks do not explicitly account for temporal relationships among events, while existing event-specific attacks employ different perturbation spaces or discrete event manipulations. This motivates an attack framework that directly optimizes point-based event inputs while incorporating both spatial and temporal neighborhood structures during adversarial generation.

%% file: sec/3_method.tex
\section{Preliminary}
\begin{figure*}
\centering
\includegraphics[width=1\linewidth]{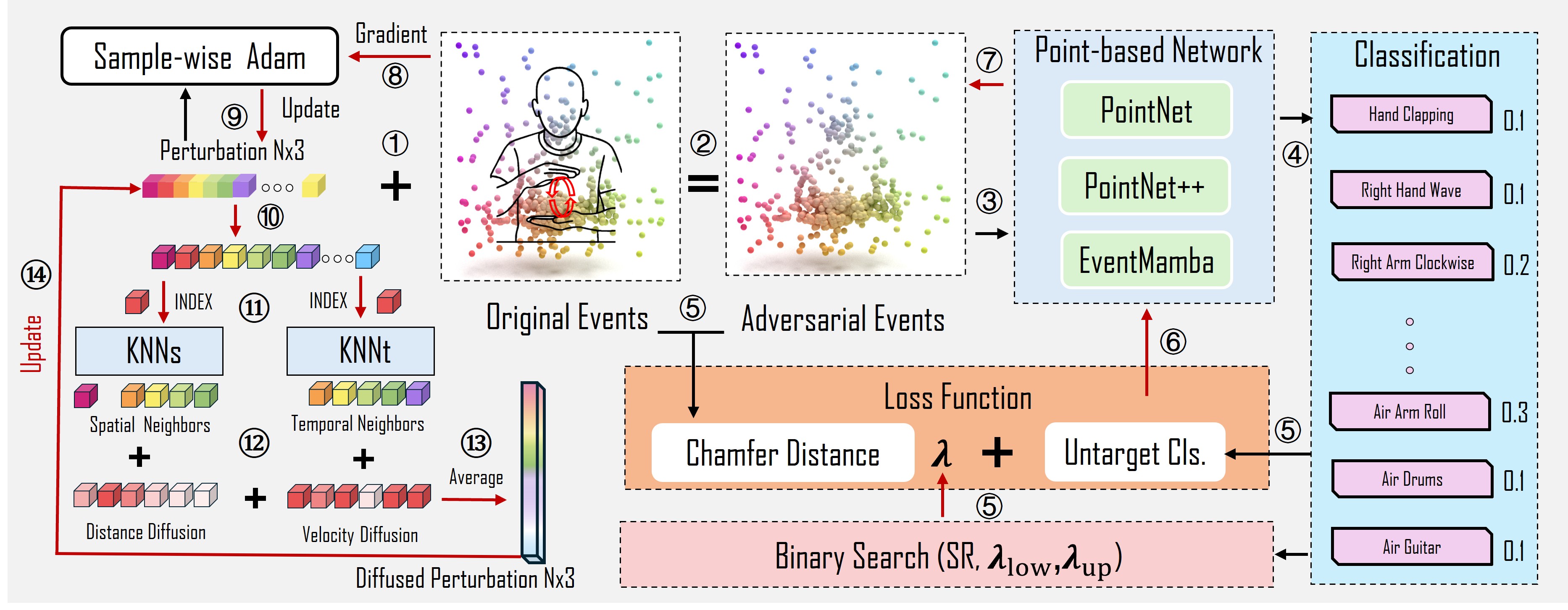}
\caption{Overview of the ST-ADV adversarial attack framework. ST-ADV generates adversarial events by incorporating spatio-temporal perturbation diffusion and sample-wise adaptive optimization. The attack objective jointly considers classification loss and perturbation cost with binary search weighting.}
\label{fig: ma-adv}
\end{figure*}

In this section, we formalize the threat model and introduce the optimization formulation for adversarial perturbations on point-based asynchronous event inputs.

\subsection{Problem Setting}

Given an asynchronous event stream
\begin{equation}
\mathcal{E}=\{e_n\}_{n=1}^{N}, \qquad
e_n=(x_n,y_n,t_n,p_n),
\end{equation}
each event contains spatial coordinates $(x_n,y_n)$, a timestamp $t_n$, and a polarity $p_n$. 
For optimization, we represent the continuous spatio-temporal coordinates as $\mathbf{X}\in\mathbb{R}^{N\times3}$, where
each row is $\mathbf{x}_n=(x_n,y_n,t_n)$. The perturbation is defined as $\mathbf{P}\in\mathbb{R}^{N\times3}$ and is
applied only to these coordinates, while the event polarities remain unchanged.
We consider a pretrained point-based event recognition model $f_{\theta}(\cdot)$ with $c$ output classes. Let $y$ denote the ground-truth class label of $\mathcal{E}$.

We study a digital white-box untargeted attack in the continuous input space of point-based event recognition models. Given a correctly classified input $\mathcal{E}$, the goal is to generate an adversarial event stream $\mathcal{E}'$ such that
\begin{equation}
\arg\max_{k} f_{\theta}(\mathcal{E}')_k \neq y,
\end{equation}
while minimizing the distortion between $\mathcal{E}$ and $\mathcal{E}'$.

The attack modifies only the continuous spatio-temporal coordinates $(x,y,t)$ of existing events, while keeping the polarity $p$ unchanged. The perturbed coordinates are constrained to the valid input ranges used by the victim model. Our threat model focuses on adversarial robustness in the continuous point-based event input space rather than sensor-level physical realizability.

\subsection{Gradient-based Adversarial Optimization}

Gradient-based adversarial attacks generate adversarial examples by optimizing perturbations through the gradients of the victim model \cite{carlini2017towards}. 
Let $\mathbf{P}$ denote the perturbation applied to the event coordinates, such that
\begin{equation}
\mathbf{X}'=\mathbf{X}+\mathbf{P},
\label{eq:perturbed_input}
\end{equation}
The adversarial event stream $\mathcal{E}'$
is then reconstructed from the perturbed
coordinates $\mathbf{X}'$ and the unchanged
polarities.

Following optimization-based adversarial attacks, we formulate the objective as
\begin{equation}
\mathcal{L}_{\mathrm{total}}
=
\mathcal{L}_{\mathrm{adv}}
+
\lambda\mathcal{L}_{\mathrm{distance}}.
\end{equation}
where $\mathcal{L}_{\mathrm{adv}}$ encourages untargeted misclassification, $\mathcal{L}_{\mathrm{distance}}$ constrains the perturbation magnitude, and $\lambda$ balances attack effectiveness and input distortion.

Following previous optimization-based
attacks~\cite{xiang2019generating},
we employ sample-wise binary search to
adjust the distance regularization weight.
Each sample maintains its own search
bounds, which are updated according to
the attack outcome:
\begin{equation}
\label{eq: binarysearch}
\lambda_{j+1}
=
\operatorname{BinSearch}
(\lambda_j,\lambda_l,\lambda_h),
\qquad
j=1,\ldots,J,
\end{equation}
where $\lambda_l$ and $\lambda_h$ denote the lower and upper search bounds, respectively, and $J$ is the total number of binary-search steps.

For each binary-search step, we initialize
the perturbation elements independently as
\begin{equation}
P_{n,d}^{(0)}
\sim\mathcal{U}(0,10^{-7}),
\label{eq:perturbation_init}
\end{equation}
where $n=1,\ldots,N$ denotes the event
index and $d\in\{1,2,3\}$ denotes the
coordinate dimension. At iteration $q$,
the perturbed input is constructed as
\begin{equation}
\mathbf{X}^{(q)}
=
\mathbf{X}+\mathbf{P}^{(q)},
\quad q=0,\ldots,I-1.
\end{equation}
The gradient with respect to the trainable
perturbation is computed as
\begin{equation}
\begin{aligned}
\mathbf{G}^{(q)}
=\nabla_{\mathbf{P}^{(q)}}\Big[
\mathcal{L}_{\mathrm{adv}}
(\mathcal{E}_{\mathrm{adv}}^{(q)},y)
+\lambda\mathcal{L}_{\mathrm{distance}}
(\mathbf{X}_{\mathrm{adv}}^{(q)},\mathbf{X})
\Big],
\end{aligned}
\label{eq:attack_gradient}
\end{equation}
where $\mathcal{E}_{\mathrm{adv}}^{(q)}$
is reconstructed from
$\mathbf{X}_{\mathrm{adv}}^{(q)}$ and the
unchanged event polarities.

The trainable perturbation is then updated
using Adam:
\begin{equation}
\mathbf{P}^{(q+1)}
=
\operatorname{AdamStep}
\left(
\mathbf{P}^{(q)},
\mathbf{G}^{(q)},
\boldsymbol{\eta}^{(q)}
\right),
\label{eq:adam_update}
\end{equation}
where $\boldsymbol{\eta}^{(q)}$ contains
the sample-specific learning rates.
The detailed update rule is presented in
the following section.

\section{Method}
In this section, we present the proposed framework for generating adversarial event streams. We first introduce the overall attack objective and then describe a spatio-temporal perturbation diffusion strategy that propagates adversarial perturbations according to local spatial and temporal relationships among asynchronous events. Finally, we present the sample-wise adaptive optimization strategy and summarize the complete attack procedure.

\subsection{Spatio-Temporal Event Perturbation Diffusion}

Unlike conventional point clouds, asynchronous event streams contain both spatial structure and explicit temporal information. 
Independently optimizing the perturbation of each event may produce locally inconsistent modifications that neglect relationships among neighboring events. 
To incorporate these structural dependencies during adversarial optimization, we construct spatial and temporal event neighborhoods and propagate perturbations among locally related events.

First, we construct two complementary neighborhoods for each event using the K-Nearest Neighbor (KNN) algorithm. 
The spatial neighborhood index $\mathcal{I}_s$ and its corresponding distance set $\mathcal{D}$ are obtained from the input event stream $\mathcal{E}$. 
In parallel, a temporal neighborhood $\mathcal{I}_t$ is constructed by restricting candidate events according to their temporal order. 
Both $\mathrm{KNN}_s$ and $\mathrm{KNN}_t$ use the joint $(x,y,t)$ distance for neighbor retrieval, while $\mathrm{KNN}_t$ only considers events occurring after the current event. 
This restriction introduces an ordered temporal dependency into perturbation propagation, as illustrated in \cref{fig: ma-adv}:
\begin{equation}
\mathcal{I}_s,\mathcal{D}
=
\mathrm{KNN}_{s}(\mathcal{E},K),
\qquad
\mathcal{I}_{t}
=
\mathrm{KNN}_{t}(\mathcal{E},K).
\end{equation}

We further introduce a lightweight spatio-temporal
variation cue to modulate perturbation propagation
within local event neighborhoods. For consecutive
events in the input sequence, we first compute
the local coordinate variation as
\begin{equation}
\mathbf{v}_i =
\begin{cases}
\operatorname{clip}\left(
\frac{\Delta\mathbf{s}_i}{\Delta t_i},
-v_{\max},v_{\max}
\right),
& |\Delta t_i|\geq\tau,\\
\mathbf{0}, & \text{otherwise}.
\end{cases}
\label{eq:local_variation}
\end{equation}
where $\mathbf{s}_i=(x_i,y_i)$ denotes the
spatial coordinates,
$\Delta\mathbf{s}_i=
\mathbf{s}_i-\mathbf{s}_{i-1}$ denotes the
spatial displacement, and
$\Delta t_i=t_i-t_{i-1}$ is the
timestamp difference.
The operator $\operatorname{clip}(\cdot,\pm v_{\max})$
restricts each component to $[-v_{\max},v_{\max}]$.
Here, $\tau$ denotes an adaptive temporal
threshold, computed as
$\tau=\max(10^{-6},0.01m_t)$, where
$m_t$ is the median of nonzero absolute
timestamp differences in the current
mini-batch. When no nonzero differences
exist, $m_t$ is set to $10^{-3}$.
The component-wise clipping bound is
defined as
$v_{\max}=\min(2Q_{0.95}/\tau,1000)$,
where $Q_{0.95}$ denotes the 95th percentile
of consecutive spatial displacement
magnitudes within the mini-batch.

The normalized variation cue is then defined as
\begin{equation}
\mathcal{C}_i =
\begin{cases}
\displaystyle
\frac{\|\mathbf{v}_i\|_2}{V_{\max}},
& V_{\max}>0,\\
0, & \text{otherwise},
\end{cases}
\label{eq:variation_cue}
\end{equation}
where $V_{\max}$ denotes the maximum variation
magnitude within the current mini-batch.
For the first event, we set
$\mathbf{v}_1=\mathbf{0}$ and $\mathcal{C}_1=0$. $\mathcal{C}_i$ is not intended to estimate physical object velocity or optical
flow. Instead, it provides a lightweight measure
of local spatio-temporal variation.
These event-wise variation cues are then used
to construct the variation matrix $\mathbf{V}$
for temporal perturbation diffusion.

Based on the spatial distances and temporal
variation cues, we define the spatial and
variation-dependent propagation weights as
\begin{equation}
\mathbf{W}_s
=\exp\left(-\frac{\mathbf{D}^{\odot 2}}{\sigma_s}\right),
\qquad
\mathbf{W}_t
=\exp\left(-\frac{\mathbf{V}}{\sigma_t}\right),
\label{eq:diffusion_weights}
\end{equation}

where $\mathbf{D}$ denotes the pairwise distance matrix
computed in the joint spatio-temporal coordinate space,
and $\mathbf{V}$ represents the temporal variation matrix.
The parameters $\sigma_s$ and $\sigma_t$ control
the decay rates of the spatial and temporal diffusion
weights, respectively.
A smaller spatial distance results in a larger
spatial propagation weight, while a smaller
variation cue increases the overall contribution
of the variation-dependent branch.
To ensure stable aggregation, the spatial and temporal weights are independently normalized within their respective neighborhoods, with a small constant $\epsilon$ added to the denominator for numerical stability.

From a graph perspective, the local event
neighborhood can be viewed as a graph, where
each event corresponds to a node and the
associated weights control the strength of
perturbation propagation between neighboring
events. Rather than explicitly solving a
graph-regularized optimization problem,
we perform lightweight weighted neighborhood
aggregation to encourage locally consistent
perturbations.

Specifically, at each optimization iteration, the current trainable perturbation is aggregated through two branches. Let \(\mathbf{p}_i\in\mathbb{R}^{3}\) denote the perturbation vector associated with event \(i\), corresponding to the \(i\)-th row of the trainable perturbation matrix \(\mathbf{P}\). 
The spatial and temporal branches propagate perturbations
over two complementary event neighborhoods. To unify
the propagation process, we formulate the final diffused
perturbation as
\begin{equation}
\tilde{\mathbf{p}}_i =
\mathbf{p}_i +
\frac{1}{2}
\sum_{r \in \{s,t\}}
\sum_{j \in I_r(i)}
\bar{w}_{ij}^{r}\mathbf{p}_j.
\label{eq:diffusion}
\end{equation}
Here, the spatial branch propagates perturbations
according to spatio-temporal proximity, whereas the
temporal branch restricts propagation to temporally
ordered neighbors. Their complementary neighborhood
structures encourage locally coherent perturbations
while preserving the temporal organization of
asynchronous events.

\subsection{Sample-wise Adaptive Optimization}

Different event samples may exhibit different optimization difficulties during adversarial generation. A fixed learning rate may therefore be insufficient for hard-to-attack samples while causing unnecessarily large updates for easier samples. To account for this variation, we introduce a sample-wise adaptive optimization strategy that adjusts the learning rate according to the attack status of each sample.

We extend Adam~\cite{kingma2014adam}
by maintaining a sample-specific learning
rate. The first- and second-moment estimates
at optimization iteration $q$ are updated as
\begin{equation}
\begin{aligned}
\mathbf{M}^{(q+1)}
&=\beta_1\mathbf{M}^{(q)}
 +(1-\beta_1)\mathbf{G}^{(q)},\\
\mathbf{V}^{(q+1)}
&=\beta_2\mathbf{V}^{(q)}
 +(1-\beta_2)(\mathbf{G}^{(q)})^2,
\end{aligned}
\label{eq:adam_moments}
\end{equation}
where $\beta_1$ and $\beta_2$ are the
momentum decay factors, and all squared
operations are performed element-wise.
Both moment estimates are initialized
to zero.
The bias-corrected moment estimates are
\begin{equation}
\widehat{\mathbf{M}}^{(q+1)}
=\frac{\mathbf{M}^{(q+1)}}
{1-\beta_1^{q+1}},
\widehat{\mathbf{V}}^{(q+1)}
=\frac{\mathbf{V}^{(q+1)}}
{1-\beta_2^{q+1}}.
\label{eq:adam_bias_correction}
\end{equation}

For each sample $k$, the learning rate
is periodically adjusted according to
its current attack status:
\begin{equation}
\eta^{(q)}[k]=
\begin{cases}
\eta^{(q-1)}[k]\rho^{(q)}[k],
& q\bmod h=0,\\
\eta^{(q-1)}[k],
& \text{otherwise},
\end{cases}
\label{eq:adaptive_lr}
\end{equation}
where
\begin{equation}
\rho^{(q)}[k]
=as^{(q)}[k]+b(1-s^{(q)}[k]).
\end{equation}
for $q\geq1$, with
$\eta^{(0)}[k]=\eta_0$.
Here, $s^{(q)}[k]\in\{0,1\}$ denotes
the attack success indicator for sample
$k$ at iteration $q$, and $h$ denotes
the adjustment interval. The factors
$a<1$ and $b>1$ decrease and increase
the learning rate for successful and
unsuccessful attacks, respectively.
The trainable perturbation is updated as
\begin{equation}
\mathbf{P}^{(q+1)}[k]
={}\mathbf{P}^{(q)}[k]
-\eta^{(q)}[k]
\frac{
\widehat{\mathbf{M}}^{(q+1)}[k]
}{
\sqrt{\widehat{\mathbf{V}}^{(q+1)}[k]}
+\epsilon
}.
\label{eq:adaptive_adam_update}
\end{equation}
where $\epsilon>0$ ensures numerical
stability, and the square root and division
are performed element-wise.

This strategy serves as an optimization refinement rather than the primary source of perturbation reduction. It decreases the update magnitude for samples that have already achieved successful attacks, while allowing harder samples to explore larger updates when necessary.

\subsection{Loss Function}

To balance attack effectiveness and perturbation distortion, we optimize a composite objective consisting of an untargeted adversarial loss and a distance regularization term.

For the untargeted attack objective, we adopt a logit-based margin loss following optimization-based adversarial attacks. The loss is defined as

\begin{equation}
\mathcal{L}_{\mathrm{adv}}
=
\frac{1}{B}
\sum_{b=1}^{B}
\max
\left(
z_b(y_b)
-
\max_{c\neq y_b} z_b(c)
+
\kappa,
0
\right),
\end{equation}
where $B$ denotes the batch size, $z_b(y_b)$ is the logit corresponding to the ground-truth class $y_b$ of the $b$-th sample, and $\max_{c\neq y_b}z_b(c)$ denotes the largest logit among the remaining classes. The margin parameter $\kappa\geq0$ controls the confidence of the untargeted attack. Minimizing this objective encourages the largest non-ground-truth logit to exceed the ground-truth logit, thereby inducing misclassification.

To constrain the perturbation magnitude, we employ a directed nearest-neighbor distance from the adversarial event set to the original event set. For each sample, the distance loss is defined as
\begin{equation}
\mathcal{L}_{\mathrm{distance}}
=
\frac{1}{N}
\sum_{i=1}^{N}
\min_{j\in\{1,\ldots,N\}}
\left\|
\mathbf{x}'_i-\mathbf{x}_j
\right\|_2,
\label{eq:distance_loss}
\end{equation}
where $\mathbf{x}'_i$ and $\mathbf{x}_j$
denote the three-dimensional spatio-temporal
coordinates of adversarial and original
events, respectively. Event polarities are
excluded from the distance computation.
The overall optimization objective is
\begin{equation}
\mathcal{L}_{\mathrm{total}}
=
\mathcal{L}_{\mathrm{adv}}(\mathcal{E}',y)
+
\lambda\mathcal{L}_{\mathrm{distance}}
(\mathbf{X}',\mathbf{X}),
\label{eq:loss_combine}
\end{equation}
where $\lambda$ controls the trade-off between attack effectiveness and perturbation distortion. Following existing optimization-based attacks, $\lambda$ is selected through the binary-search procedure described in \cref{eq: binarysearch}.

%% file: sec/4_experiment.tex
\section{Experiments}
In this section, we systematically validate the effectiveness and robustness of the proposed ST-ADV framework through comprehensive experiments. The platform configuration comprises the following: CPU:
AMD 7950x, GPU: RTX 4090, and Memory: 32GB.
\begin{table*}[!t]
\caption{Performance metrics of three models under six adversarial attack methods. Specifically, the abbreviation "SR" denotes Success Rate, \(\mathbb{D}_{\text{chamfer}}\) represents Chamfer Distance, \(\mathbb{D}_{L_2}\) is short for $L_2$ Distance, and \(\mathbb{D}_{\text{hausdorff}}\) signifies Hausdorff Distance.}
\centering
\renewcommand{\arraystretch}{1.1}
\resizebox{\linewidth}{!}{%
\begin{tabular}{cl*{12}{c}}
\toprule
\multirow{2}{*}{Dataset} & \multirow{2}{*}{Method} & \multicolumn{4}{c}{\textbf{EventMamba}} & \multicolumn{4}{c}{\textbf{PointNet++}} & \multicolumn{4}{c}{\textbf{PointNet}} \\
\cmidrule(lr){3-6} \cmidrule(lr){7-10} \cmidrule(lr){11-14}
& & SR $\uparrow$ & $\mathbb{D}_{\mathrm{chamfer}}\downarrow$ & $\mathbb{D}_{L_2}\downarrow$ & $\mathbb{D}_{\mathrm{hausdorff}}\downarrow$ & SR $\uparrow$ & $\mathbb{D}_{\mathrm{chamfer}}\downarrow$ & $\mathbb{D}_{L_2}\downarrow$ & $\mathbb{D}_{\mathrm{hausdorff}}\downarrow$ & SR $\uparrow$ & $\mathbb{D}_{\mathrm{chamfer}}\downarrow$ & $\mathbb{D}_{L_2}\downarrow$ & $\mathbb{D}_{\mathrm{hausdorff}}\downarrow$ \\
\midrule
\multirow{6}{*}{\rotatebox{90}{DVSGesture}} & FGSM & 0.9414 & 0.43360055 & 44.514780 & 0.83827629 & 0.9416 & 0.03886039 & 43.480246 & 0.37155187 & 0.8927 & 0.00634454 & 28.684870 & 0.23422670 \\
 & IFGSM & 0.8861 & 0.00406675 & 6.107059 & 0.12109839 & 0.9821 & 0.00274648 & 5.045233 & 0.06968748 & 0.9964 & 0.00220336 & 2.767722 & 0.06962123 \\
 & C\&W+IFGSM & \textbf{1.0000} & 0.03150661 & 10.622097 & 0.20722541 & \textbf{1.0000} & 0.01487889 & 5.816218 & 0.09292601 & \textbf{1.0000} & 0.00126929 & 1.269730 & 0.00995742 \\
 & HiT-ADV & 0.9138 & 0.01826267 & 8.495867 & 0.09952199 & 0.3485 & 0.00238013 & 1.368787 & 0.01647898 & \textbf{1.0000} & 0.00114099 & 1.541720 & 0.00323261 \\
 & C\&W & 0.9977 & 0.00064274 & 1.127571 & 0.00840837 & \textbf{1.0000} & 0.00028885 & 0.692771 & 0.00386352 & 0.9997 & 0.00001346 & 0.093563 & 0.00137043 \\

 \rowcolor{gray!20} & ST-ADV   & \textbf{1.0000} & \textbf{0.00057679} & \textbf{1.019651} & \textbf{0.00729263} & \textbf{1.0000} & \textbf{0.00010291} & \textbf{0.346385} & \textbf{0.00221654} & \textbf{1.0000} & \textbf{0.00001106} & \textbf{0.088857} & \textbf{0.00119464} \\
\midrule
\multirow{6}{*}{\rotatebox{90}{N-Caltech101}} & FGSM & 0.9842 & 0.00660517 & 117.385498 & 0.14525289 & 0.7101 & 0.00225719 & 104.926770 & 0.09454342 & 0.9966 & 0.00019279 & 45.315790 & 0.03437342 \\
 & IFGSM & \textbf{1.0000} & 0.00112684 & 14.133716 & 0.03765472 & \textbf{1.0000} & 0.00078093 & 10.291948 & 0.02683330 & 0.9971 & 0.00053478 & 4.365599 & 0.02439225 \\
 & C\&W+IFGSM & \textbf{1.0000} & 0.00013692 & 0.832648 & 0.00139306 & \textbf{1.0000} & 0.00060845 & 3.760537 & 0.00917240 & \textbf{1.0000} & 0.00008581 & 0.933215 & 0.00097653 \\
 & HiT-ADV & \textbf{1.0000} & 0.00020516 & 1.243481 & 0.00115040 & 0.8053 & 0.00794351 & 8.289549 & 0.04941864 & \textbf{1.0000} & 0.00276941 & 11.064562 & 0.02378495 \\
 & C\&W & \textbf{1.0000} & 0.00003940 & 0.343619 & \textbf{0.00054358} & \textbf{1.0000} & 0.00010992 & 1.007796 & 0.00375492 & \textbf{1.0000} & 0.00000517 & 0.154703 & 0.00109625 \\

 \rowcolor{gray!20} & ST-ADV   & \textbf{1.0000} & \textbf{0.00002142} & \textbf{0.242915} & 0.00116328 & \textbf{1.0000} & \textbf{0.00003151} & \textbf{0.442692} & \textbf{0.00339322} & \textbf{1.0000} & \textbf{0.00000118} & \textbf{0.075429} & \textbf{0.00088142} \\
\midrule
\multirow{6}{*}{\rotatebox{90}{N-MNIST}} & FGSM & 0.8754 & 0.02763776 & 88.102110 & 0.29325186 & 0.5592 & 0.00173537 & 65.676482 & 0.13920432 & 0.8962 & 0.00053466 & 38.446476 & 0.12184391 \\
 & IFGSM & 0.9133 & 0.00149740 & 10.990860 & 0.06823992 & 0.9999 & 0.00100511 & 7.642727 & 0.05615357 & 0.9556 & 0.00081456 & 3.946213 & 0.06157463 \\
 & C\&W+IFGSM & \textbf{1.0000} & 0.01881020 & 20.719540 & 0.15299544 & \textbf{1.0000} & 0.01559783 & 22.652911 & 0.19563197 & \textbf{1.0000} & 0.00095307 & 5.574319 & 0.02148315 \\
 & HiT-ADV & 0.9984 & 0.02081660 & 22.843958 & 0.15220663 & 0.2733 & 0.00374189 & 4.696957 & 0.04063487 & \textbf{1.0000} & 0.00239242 & 6.587781 & 0.01286503 \\
 & C\&W & \textbf{1.0000} & 0.00035671 & 2.282792 & 0.00683590 & \textbf{1.0000} & 0.00024479 & 1.650611 & 0.00650290 & \textbf{1.0000} & 0.00000690 & 0.195035 & 0.00135394 \\

 \rowcolor{gray!20} & ST-ADV  & \textbf{1.0000} & \textbf{0.00033453} & \textbf{2.049050} & \textbf{0.00654472} & \textbf{1.0000} & \textbf{0.00012078} & \textbf{0.932534} & \textbf{0.00465204} & \textbf{1.0000} & \textbf{0.00000284} & \textbf{0.119965} & \textbf{0.00090106} \\
\bottomrule
\end{tabular}%
}
\label{tab:attack_performance_complete}
\label{table: main results}
\end{table*}

\subsection{Experimental Setup}
Our experiments are conducted on three public event-based datasets, including DVSGesture \cite{amir2017low}, NCALTECH \cite{orchard2015converting}, and N-MNIST \cite{orchard2015converting}, covering dynamic gesture recognition, object recognition, and handwritten digit classification scenarios. These datasets provide diverse event streams with different temporal dynamics, spatial structures, and class variations.

We evaluate ST-ADV on three representative event-based recognition models, including EventMamba \cite{ren2025rethinking}, PointNet++ \cite{qi2017pointnet++}, and PointNet \cite{qi2017pointnet}. For models with available pretrained weights, we directly use the official checkpoints; otherwise, we train the models on the corresponding datasets and use the obtained models as attack targets.

\subsection{Implementation Details}
The total number of optimization iterations is set to $I=100$, and the binary search step is set to $J=20$. The initial learning rate is set to $\eta=1e{-2}$, which is sufficient to ensure stable optimization while maintaining efficient convergence. For the binary-search-based loss balancing strategy, we set the lower and upper bounds as $\lambda_{l}=0$ and $\lambda_{h}=80$, respectively, so that the attack objective can be adaptively adjusted within a reasonable range.

For the perturbation diffusion module, the KNN number is set to $K=10$ to capture local spatio-temporal neighbors for adversarial propagation. The diffusion weights are controlled by $\sigma_{s}=0.01$ and $\sigma_{t}=0.1$, which determine the influence of spatial and temporal distances, respectively. 
The parameters $a=0.8$ and $b=1.2$
control the sample-wise learning rate
adjustment for successful and unsuccessful
attacks, respectively. Unless otherwise noted, these hyperparameters are fixed across all experiments.

\subsection{Evaluation Metrics}
We assess the effectiveness of adversarial attacks using the attack Success Rate (SR), which is defined as the percentage of generated adversarial examples that successfully mislead the target classifiers. A higher SR indicates that the attack is more effective in causing misclassification, therefore, better exposes the vulnerability of the victim model.

To measure the distortion cost of the generated adversarial events, we further adopt three distance metrics: Chamfer distance \cite{borgefors1986distance}, Hausdorff distance \cite{huttenlocher1993comparing}, and the $L_2$ norm distance \cite{goodfellow2014explaining}. 

To further evaluate whether adversarial perturbations preserve event-specific spatio-temporal structures, we introduce two event-aware metrics: Normalized Temporal Perturbation (NTP) and Temporal Surface Consistency (TSC).
1. Normalized Temporal Perturbation (NTP) measures the average timestamp deviation between the original and adversarial event streams, normalized by the temporal duration of the original stream. Lower NTP values indicate smaller temporal distortion:
\begin{equation}
\mathrm{NTP}(\mathcal{E}, \mathcal{E}')
=
\frac{1}{N}
\sum_{i=1}^{N}
\frac{|t'_i-t_i|}
{t_{\max}-t_{\min}+\epsilon}.
\end{equation}
2. Temporal Surface Consistency (TSC) measures the difference between the time surfaces of the original and adversarial event streams, evaluating the preservation of local spatio-temporal event structures. Lower TSC values indicate stronger temporal consistency:
\begin{equation}
\mathrm{TSC}(\mathcal{E}, \mathcal{E}')
=
\frac{1}{|\Omega^+|}
\sum_{(u,v)\in\Omega^+}
\left|
S_{\mathcal{E}}(u,v)-S_{\mathcal{E}'}(u,v)
\right|.
\end{equation}

\subsection{Attack and Defense Methods}
We compare ST-ADV with five representative point-cloud adversarial attacks, including FGSM \cite{goodfellow2014explaining}, IFGSM \cite{liu2019extending}, C\&W + IFGSM \cite{xiang2019generating}, C\&W \cite{xiang2019generating}, and HIT-ADV \cite{lou2024hide}. These baselines cover gradient-based, iterative optimization-based, and geometry-aware attack strategies, providing diverse references for evaluating attack effectiveness and perturbation efficiency. Since point-based event models process raw events at the individual event level, these attacks serve as relevant baselines for comparison.

For defense evaluation, we adopt three representative point-cloud defenses, including SOR \cite{rusu20113d}, SRS \cite{lou2024hide}, and DUPNet \cite{zhou2019dup}, which represent statistical filtering, random sampling, and denoising-based purification strategies, respectively. These defenses allow us to evaluate the robustness of ST-ADV under different input preprocessing mechanisms.

\subsection{Attack Results}
\begin{table}[t]
\caption{Event-specific perturbation analysis of C\&W and ST-ADV on different event-based recognition models.}
\centering
\scalebox{0.8}{
\begin{tabular}{clcccccc}
\toprule
\multirow{2}{*}{Method}
& \multicolumn{2}{c}{\textbf{DVSGesture}}
& \multicolumn{2}{c}{\textbf{NCALTECH}}
& \multicolumn{2}{c}{\textbf{N-MNIST}}\\
\cmidrule(lr){2-3}
\cmidrule(lr){4-5}
\cmidrule(lr){6-7}
& NTP $\downarrow$ & TSC $\downarrow$
& NTP $\downarrow$ & TSC $\downarrow$
& NTP $\downarrow$ & TSC $\downarrow$\\
\midrule

\multicolumn{7}{c}{\textbf{PointNet}}\\
\midrule
C\&W
&0.0007&0.1135
&0.0003&0.0501
&0.0006&0.1974\\

\rowcolor{gray!20}
ST-ADV
&\textbf{0.0006}&\textbf{0.1038}
&\textbf{0.0001}&\textbf{0.0245}
&\textbf{0.0003}&\textbf{0.1468}\\

\midrule

\multicolumn{7}{c}{\textbf{PointNet++}}\\
\midrule
C\&W
&0.0095&0.3335
&0.0030&0.2014
&0.0091&0.4731\\

\rowcolor{gray!20}
ST-ADV
&\textbf{0.0043}&\textbf{0.2128}
&\textbf{0.0009}&\textbf{0.1410}
&\textbf{0.0040}&\textbf{0.3553}\\

\midrule

\multicolumn{7}{c}{\textbf{EventMamba}}\\
\midrule
C\&W
&0.0154&0.3764
&0.0016&0.0901
&\textbf{0.0124}&0.4902\\

\rowcolor{gray!20}
ST-ADV
&\textbf{0.0138}&\textbf{0.3627}
&\textbf{0.0008}&\textbf{0.0672}
&0.0127&\textbf{0.4898}\\

\bottomrule
\end{tabular}
}
\label{table:event metric}
\end{table}

\begin{figure*}
\centering
\includegraphics[width=1\linewidth]{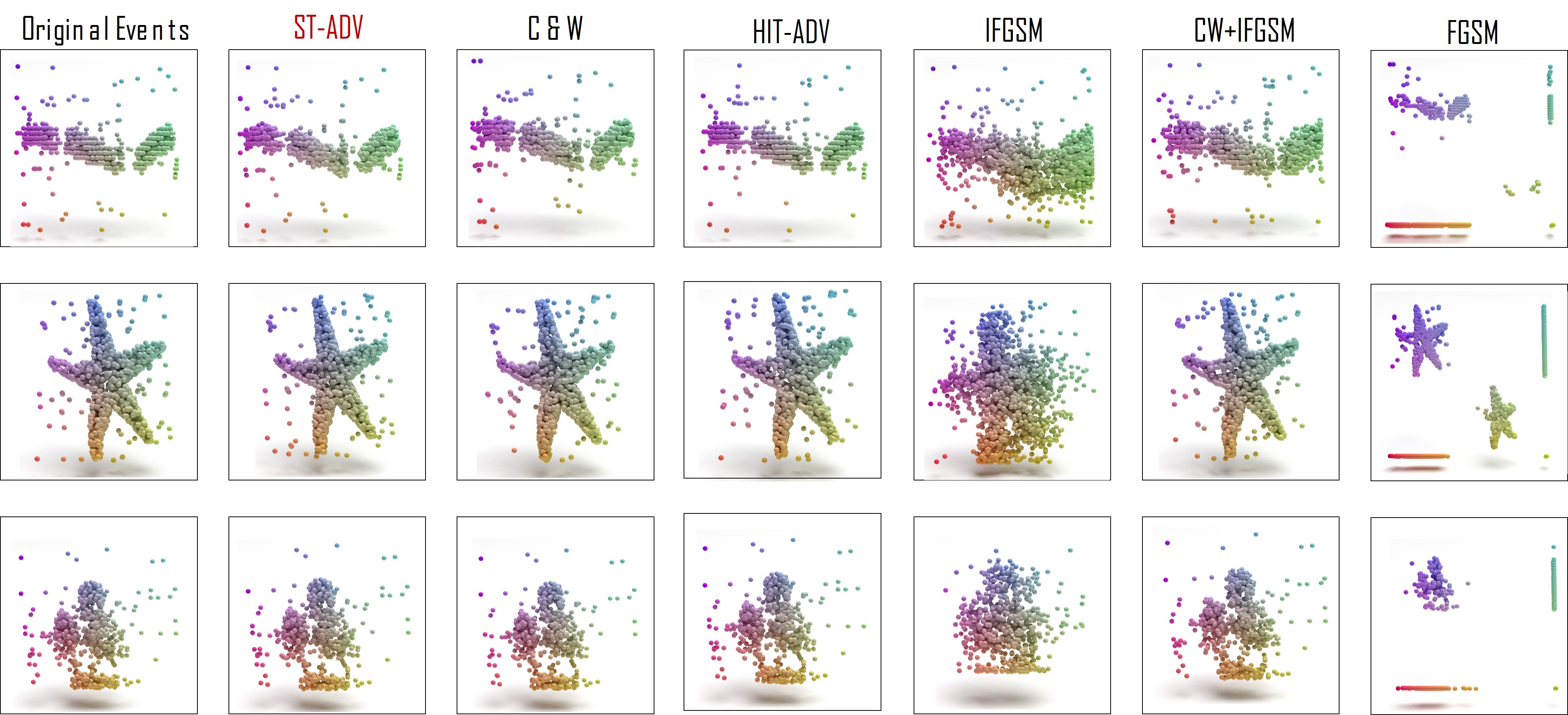}
\caption{Visualization of adversarial event samples across different attack methods on three datasets. Zoom in to find details.}
\label{fig: visualization}
\end{figure*}
\cref{table: main results} summarizes the evaluation results of ST-ADV and six adversarial attacks across three point-based backbones and three event datasets. Overall, gradient-based attacks usually require larger perturbations to achieve high attack success rates, while optimization-based methods provide a better trade-off between attack effectiveness and distortion.
Among all baselines, ST-ADV achieves a 100\% attack success rate across all evaluated settings, consistently yielding the lowest Chamfer and L2 distances while maintaining competitive Hausdorff distances.
These results demonstrate that the proposed spatio-temporal perturbation diffusion can generate effective adversarial events with reduced perturbation costs.

The results also reveal different robustness levels among victim models. PointNet is generally more vulnerable to adversarial perturbations, whereas EventMamba requires larger perturbations to achieve successful attacks, indicating that different architectures exhibit varying robustness to adversarial event streams. Nevertheless, ST-ADV maintains consistent advantages across datasets and backbones, demonstrating its effectiveness and generality. We further visualize adversarial examples in \cref{fig: visualization} and compare with existing event-specific attacks \cite{lee2022adversarial,buchel2022adversarial}. Although previous methods report competitive attack success rates, they mainly focus on specific event manipulation strategies and lack distance-based perturbation evaluation. In contrast, ST-ADV achieves comparable or superior attack effectiveness while additionally providing quantitative analysis of perturbation quality. 

However, conventional geometric distances cannot fully characterize distortions in asynchronous event streams. Beyond conventional distortion metrics, we further evaluate event-specific perturbation quality using NTP and TSC, as shown in \cref{table:event metric}. ST-ADV generally achieves lower temporal distortion and time-surface differences than C\&W across different settings, indicating better preservation of event spatio-temporal structures. These results demonstrate that the proposed spatio-temporal diffusion strategy improves attack effectiveness while maintaining event-level consistency.

\subsection{Defense Results}
\begin{table}[t]
\centering
\caption{Attack success rate (SR, $\uparrow$) of EventMamba under different defense strategies on three datasets. The best results are highlighted in bold.}
\label{tab:defense_results}
\footnotesize
\setlength{\tabcolsep}{3pt}
\renewcommand{\arraystretch}{1.15}
\begin{tabular}{@{}llcccc@{}}
\toprule
Dataset & Attack & No Defense & SOR & SRS & DUP-Net \\
\midrule
\multirow{2}{*}{DVSGesture}
& C\&W   & 0.9977 & 0.5442 & 0.4128 & \textbf{0.4183} \\
& ST-ADV & \textbf{1.0000} & \textbf{0.6187} & \textbf{0.4277} & 0.4108 \\
\midrule
\multirow{2}{*}{N-Caltech101}
& C\&W   & \textbf{1.0000} & 0.3020 & 0.1531 & 0.3004 \\
& ST-ADV & \textbf{1.0000} & \textbf{0.3709} & \textbf{0.1638} & \textbf{0.3145} \\
\midrule
\multirow{2}{*}{N-MNIST}
& C\&W   & \textbf{1.0000} & 0.2864 & 0.4259 & 0.4333 \\
& ST-ADV & \textbf{1.0000} & \textbf{0.4598} & \textbf{0.4701} & \textbf{0.4688} \\
\bottomrule
\end{tabular}
\label{tab: defend method}
\end{table}

\cref{tab: defend method} presents the performance of EventMamba under different defense strategies across three datasets. Although all defenses reduce the attack success rates compared with the undefended setting, ST-ADV consistently maintains higher or comparable attack effectiveness than C\&W under most defense scenarios. This indicates that the adversarial events generated by ST-ADV are more robust to common preprocessing-based defenses. A possible explanation is that ST-ADV introduces perturbations by exploiting local spatio-temporal relationships among events rather than isolated irregular changes, allowing adversarial effects to remain after partial removal or filtering. These results demonstrate that evaluating attacks under defended settings provides a more comprehensive assessment of adversarial robustness for event-based recognition models.

\subsection{Ablation Study}
\begin{table}[!t]
\caption{Ablation study on different module configurations.}
\centering
\resizebox{\linewidth}{!}{%
\begin{tabular}{lcccc}
\toprule
Condition
& SR $\uparrow$
& $\mathbb{D}_{\mathrm{chamfer}}$ $(\times 10^{-4})$ $\downarrow$
& $\mathbb{D}_{L_2}$ $\downarrow$
& $\mathbb{D}_{\mathrm{hausdorff}}$ $(\times 10^{-3})$ $\downarrow$ \\
\midrule
w/o Temporal    & \textbf{1.0000} & 5.9670 & 1.0405 & 7.6755 \\
w/o Spatial     & \textbf{1.0000} & 5.8204 & 1.0589 & 7.1790 \\
w/o Diffusion   & 0.9986          & 6.3898 & 1.1579 & 8.6451 \\
w/o Adaptive lr & 0.9997          & 5.9106 & 1.0326 & \textbf{7.0510} \\
w/o Causal      & \textbf{1.0000} & 5.8744 & 1.0356 & 7.3145 \\
\rowcolor{gray!20}
ST-ADV          & \textbf{1.0000} & \textbf{5.7679}
                & \textbf{1.0197} & 7.2926 \\
\bottomrule
\end{tabular}%
}
\label{tab: module ablation}
\end{table}

\textbf{Module Ablation:}
\cref{tab: module ablation} evaluates the contributions
of individual components in ST-ADV. All variants maintain
high attack success rates, while removing either the
spatial or temporal diffusion branch increases both
Chamfer and L2 distances, indicating their complementary
roles in reducing perturbation costs. Disabling the
entire diffusion module further increases Chamfer,
L2, and Hausdorff distances, highlighting the importance
of neighborhood-based perturbation propagation.
Removing the adaptive learning rate or temporal
ordering results in relatively modest increases in
Chamfer and L2 distances, although their effects
on Hausdorff distance are mixed. Overall, the complete
ST-ADV achieves the lowest Chamfer and L2 distances
while maintaining a 100\% attack success rate,
demonstrating the benefits of combining spatio-temporal
diffusion with adaptive optimization.
\begin{table}[!t]
\caption{Ablation study on the number of diffusion points.
}
\centering
\resizebox{\linewidth}{!}{%
\begin{tabular}{ccccc}
\toprule
Number & SR $\uparrow$ & $\mathbb{D}_{\mathrm{chamfer}}$ $(\times10^{-4})\downarrow$ & $\mathbb{D}_{L_2}\downarrow$ & $\mathbb{D}_{\mathrm{hausdorff}}$ $(\times10^{-3})\downarrow$ \\
\midrule
5 & 1.0000 & 6.1477 & 1.0733 & 7.7858 \\
\rowcolor{gray!20}
10 & 1.0000 & 5.7679 & 1.0197 & 7.2926 \\
15 & 1.0000 & 5.8111 & 1.0333 & 7.2370 \\
20 & 1.0000 & 5.8623 & 1.0256 & 7.2325 \\
25 & 0.9998 & 5.7695 & 1.0279 & 7.2290 \\
\bottomrule
\end{tabular}%
}
\label{tab: diffusion points}
\end{table}

\textbf{Diffusion Points Number Ablation:} 
\cref{tab: diffusion points} investigates the influence of the number of diffusion points on ST-ADV. The results show that ST-ADV remains stable across different settings, achieving consistently high attack success rates. 
Using 10 diffusion points achieves the
lowest Chamfer and L2 distances while
maintaining a 100\% attack success rate,
although larger neighborhoods yield
slightly lower Hausdorff distances.
A smaller number of diffusion points may limit the exploitation of local spatio-temporal relationships, whereas excessive diffusion may introduce unnecessary perturbation propagation and weaken local discriminative effects. Therefore, we adopt 10 diffusion points as the default setting.

\section{Conclusion}
This paper proposes ST-ADV, a spatio-temporal adversarial framework for generating adversarial events via point cloud representations. By introducing a spatio-temporal perturbation diffusion strategy and a sample-wise learning rate adjustment mechanism, ST-ADV effectively mitigates high-frequency event noise and optimizes perturbation cost. Extensive experiments demonstrate the effectiveness and robustness of ST-ADV.

\section{LLM Usage Statement}
OpenAI's ChatGPT (GPT models) was used to assist with editing and refining author-produced content throughout the manuscript, including language polishing, clarity, and presentation. All AI-assisted revisions were reviewed by the authors, who take full responsibility for the scientific content, accuracy, and integrity of the submitted work.

%% file: main.bib
@String(ICIP = {IEEE Int. Conf. Image Process.})

@String(AAAI = {AAAI})

@String(ICIP  = {ICIP})

@article{lecun2015deep,
  title={Deep learning},
  author={LeCun, Yann and Bengio, Yoshua and Hinton, Geoffrey},
  journal={nature},
  volume={521},
  number={7553},
  pages={436--444},
  year={2015},
  publisher={Nature Publishing Group UK London}
}

@inproceedings{xiang2019generating,
  title={Generating 3d adversarial point clouds},
  author={Xiang, Chong and Qi, Charles R and Li, Bo},
  booktitle={Proceedings of the IEEE/CVF conference on computer vision and pattern recognition},
  pages={9136--9144},
  year={2019}
}

@article{zhang2025adversarial,
  title={Adversarial attacks of vision tasks in the past 10 years: A survey},
  author={Zhang, Chiyu and Zhou, Lu and Xu, Xiaogang and Wu, Jiafei and Liu, Zhe},
  journal={ACM Computing Surveys},
  volume={58},
  number={2},
  pages={1--42},
  year={2025},
  publisher={ACM New York, NY}
}

@article{gallego2020event,
  title={Event-based vision: A survey},
  author={Gallego, Guillermo and Delbr{\"u}ck, Tobi and Orchard, Garrick and Bartolozzi, Chiara and Taba, Brian and Censi, Andrea and Leutenegger, Stefan and Davison, Andrew J and Conradt, J{\"o}rg and Daniilidis, Kostas and others},
  journal={IEEE transactions on pattern analysis and machine intelligence},
  volume={44},
  number={1},
  pages={154--180},
  year={2020},
  publisher={IEEE}
}

@article{zhu2018ev,
  title={EV-FlowNet: Self-supervised optical flow estimation for event-based cameras},
  author={Zhu, Alex Zihao and Yuan, Liangzhe and Chaney, Kenneth and Daniilidis, Kostas},
  journal={arXiv preprint arXiv:1802.06898},
  year={2018}
}

@incollection{kurakin2018adversarial,
  title={Adversarial examples in the physical world},
  author={Kurakin, Alexey and Goodfellow, Ian J and Bengio, Samy},
  booktitle={Artificial intelligence safety and security},
  pages={99--112},
  year={2018},
  publisher={Chapman and Hall/CRC}
}

@article{goodfellow2014explaining,
  title={Explaining and harnessing adversarial examples},
  author={Goodfellow, Ian J and Shlens, Jonathon and Szegedy, Christian},
  journal={arXiv preprint arXiv:1412.6572},
  year={2014}
}

@inproceedings{qi2017pointnet,
  title={Pointnet: Deep learning on point sets for 3d classification and segmentation},
  author={Qi, Charles R and Su, Hao and Mo, Kaichun and Guibas, Leonidas J},
  booktitle={Proceedings of the IEEE conference on computer vision and pattern recognition},
  pages={652--660},
  year={2017}
}

@article{ren2025rethinking,
  title={Rethinking efficient and effective point-based networks for event camera classification and regression},
  author={Ren, Hongwei and Zhou, Yue and Zhu, Jiadong and Lin, Xiaopeng and Fu, Haotian and Huang, Yulong and Fang, Yuetong and Ma, Fei and Yu, Hao and Cheng, Bojun},
  journal={IEEE Transactions on Pattern Analysis and Machine Intelligence},
  year={2025},
  publisher={IEEE}
}

@article{wen2020geometry,
  title={Geometry-aware generation of adversarial point clouds},
  author={Wen, Yuxin and Lin, Jiehong and Chen, Ke and Chen, CL Philip and Jia, Kui},
  journal={IEEE Transactions on Pattern Analysis and Machine Intelligence},
  volume={44},
  number={6},
  pages={2984--2999},
  year={2020},
  publisher={IEEE}
}

@article{kingma2014adam,
  title={Adam: A method for stochastic optimization},
  author={Kingma, Diederik P and Ba, Jimmy},
  journal={arXiv preprint arXiv:1412.6980},
  year={2014}
}

@inproceedings{amir2017low,
  title={A low power, fully event-based gesture recognition system},
  author={Amir, Arnon and Taba, Brian and Berg, David and Melano, Timothy and McKinstry, Jeffrey and Di Nolfo, Carmelo and Nayak, Tapan and Andreopoulos, Alexander and Garreau, Guillaume and Mendoza, Marcela and others},
  booktitle={Proceedings of the IEEE conference on computer vision and pattern recognition},
  pages={7243--7252},
  year={2017}
}

@inproceedings{rusu20113d,
  title={3d is here: Point cloud library (pcl)},
  author={Rusu, Radu Bogdan and Cousins, Steve},
  booktitle={2011 IEEE international conference on robotics and automation},
  pages={1--4},
  year={2011},
  organization={IEEE}
}

@inproceedings{zhou2019dup,
  title={Dup-net: Denoiser and upsampler network for 3d adversarial point clouds defense},
  author={Zhou, Hang and Chen, Kejiang and Zhang, Weiming and Fang, Han and Zhou, Wenbo and Yu, Nenghai},
  booktitle={Proceedings of the IEEE/CVF international conference on computer vision},
  pages={1961--1970},
  year={2019}
}

@article{huttenlocher1993comparing,
  title={Comparing images using the Hausdorff distance},
  author={Huttenlocher, Daniel P and Klanderman, Gregory A and Rucklidge, William and others},
  journal={IEEE Transactions on pattern analysis and machine intelligence},
  volume={15},
  number={9},
  pages={850--863},
  year={1993}
}

@article{borgefors1986distance,
  title={Distance transformations in digital images},
  author={Borgefors, Gunilla},
  journal={Computer vision, graphics, and image processing},
  volume={34},
  number={3},
  pages={344--371},
  year={1986},
  publisher={Elsevier}
}

@article{orchard2015converting,
  title={Converting static image datasets to spiking neuromorphic datasets using saccades},
  author={Orchard, Garrick and Jayawant, Ajinkya and Cohen, Gregory K and Thakor, Nitish},
  journal={Frontiers in neuroscience},
  volume={9},
  pages={437},
  year={2015},
  publisher={Frontiers Media SA}
}

@article{qi2017pointnet++,
  title={Pointnet++: Deep hierarchical feature learning on point sets in a metric space},
  author={Qi, Charles Ruizhongtai and Yi, Li and Su, Hao and Guibas, Leonidas J},
  journal={Advances in neural information processing systems},
  volume={30},
  year={2017}
}

@inproceedings{lou2024hide,
  title={Hide in thicket: Generating imperceptible and rational adversarial perturbations on 3d point clouds},
  author={Lou, Tianrui and Jia, Xiaojun and Gu, Jindong and Liu, Li and Liang, Siyuan and He, Bangyan and Cao, Xiaochun},
  booktitle={Proceedings of the IEEE/CVF Conference on Computer Vision and Pattern Recognition},
  pages={24326--24335},
  year={2024}
}

@article{buchel2022adversarial,
  title={Adversarial attacks on spiking convolutional neural networks for event-based vision},
  author={B{\"u}chel, Julian and Lenz, Gregor and Hu, Yalun and Sheik, Sadique and Sorbaro, Martino},
  journal={Frontiers in Neuroscience},
  volume={16},
  pages={1068193},
  year={2022},
  publisher={Frontiers Media SA}
}

@inproceedings{lin2025adversarial,
  title={Adversarial Attacks on Event-Based Pedestrian Detectors: A Physical Approach},
  author={Lin, Guixu and Niu, Muyao and Zhu, Qingtian and Yin, Zhengwei and Li, Zhuoxiao and He, Shengfeng and Zheng, Yinqiang},
  booktitle={Proceedings of the AAAI Conference on Artificial Intelligence},
  volume={39},
  number={5},
  pages={5227--5235},
  year={2025}
}

@inproceedings{lee2022adversarial,
  title={Adversarial attack for asynchronous event-based data},
  author={Lee, Wooju and Myung, Hyun},
  booktitle={Proceedings of the AAAI Conference on Artificial Intelligence},
  volume={36},
  number={2},
  pages={1237--1244},
  year={2022}
}

@article{wei2025smokeattack,
  title={SmokeAttack: Physically-Based Adversarial Smoke for LiDAR Point Cloud Detectors},
  author={Wei, Xuqin and Zheng, Shijun and Yang, Lina},
  journal={Pattern Recognition},
  pages={112920},
  year={2025},
  publisher={Elsevier}
}

@inproceedings{tang2023deep,
  title={Deep manifold attack on point clouds via parameter plane stretching},
  author={Tang, Keke and Wu, Jianpeng and Peng, Weilong and Shi, Yawen and Song, Peng and Gu, Zhaoquan and Tian, Zhihong and Wang, Wenping},
  booktitle={Proceedings of the AAAI Conference on Artificial Intelligence},
  volume={37},
  number={2},
  pages={2420--2428},
  year={2023}
}

@inproceedings{huang2022shape,
  title={Shape-invariant 3D adversarial point clouds},
  author={Huang, Qidong and Dong, Xiaoyi and Chen, Dongdong and Zhou, Hang and Zhang, Weiming and Yu, Nenghai},
  booktitle={Proceedings of the IEEE/CVF conference on computer vision and pattern recognition},
  pages={15335--15344},
  year={2022}
}

@inproceedings{liu2019extending,
  title={Extending adversarial attacks and defenses to deep 3d point cloud classifiers},
  author={Liu, Daniel and Yu, Ronald and Su, Hao},
  booktitle={2019 IEEE International Conference on Image Processing (ICIP)},
  pages={2279--2283},
  year={2019},
  organization={IEEE}
}

@inproceedings{ren2024spikepoint,
  title={Spikepoint: An efficient point-based spiking neural network for event cameras action recognition},
  author={Ren, Hongwei and Zhou, Yue and Lin, Xiaopeng and Huang, Yulong and Fu, Haotian and Song, Jie and Cheng, Bojun},
  booktitle={International Conference on Learning Representations},
  volume={2024},
  pages={27827--27846},
  year={2024}
}

@inproceedings{yao2024exploring,
  title={Exploring vulnerabilities in spiking neural networks: Direct adversarial attacks on raw event data},
  author={Yao, Yanmeng and Zhao, Xiaohan and Gu, Bin},
  booktitle={European conference on computer vision},
  pages={412--428},
  year={2024},
  organization={Springer}
}

@inproceedings{carlini2017towards,
  title={Towards evaluating the robustness of neural networks},
  author={Carlini, Nicholas and Wagner, David},
  booktitle={2017 ieee symposium on security and privacy (sp)},
  pages={39--57},
  year={2017},
  organization={Ieee}
}

@inproceedings{kim2021minimal,
  title={Minimal adversarial examples for deep learning on 3d point clouds},
  author={Kim, Jaeyeon and Hua, Binh-Son and Nguyen, Thanh and Yeung, Sai-Kit},
  booktitle={Proceedings of the IEEE/CVF international conference on computer vision},
  pages={7797--7806},
  year={2021}
}

@inproceedings{hamdi2020advpc,
  title={Advpc: Transferable adversarial perturbations on 3d point clouds},
  author={Hamdi, Abdullah and Rojas, Sara and Thabet, Ali and Ghanem, Bernard},
  booktitle={European conference on computer vision},
  pages={241--257},
  year={2020},
  organization={Springer}
}
